\documentclass[letterpaper]{article} % DO NOT CHANGE THIS
\usepackage{aaai25}  % DO NOT CHANGE THIS
\usepackage{times}  % DO NOT CHANGE THIS
\usepackage{helvet}  % DO NOT CHANGE THIS
\usepackage{courier}  % DO NOT CHANGE THIS
\usepackage[hyphens]{url}  % DO NOT CHANGE THIS
\usepackage{graphicx} % DO NOT CHANGE THIS
\usepackage{natbib}  % DO NOT CHANGE THIS AND DO NOT ADD ANY OPTIONS TO IT
\usepackage{caption} % DO NOT CHANGE THIS AND DO NOT ADD ANY OPTIONS TO IT
\usepackage{algorithm}
\usepackage{algorithmic}

\usepackage{newfloat}
\usepackage{listings}
\DeclareCaptionStyle{ruled}{labelfont=normalfont,labelsep=colon,strut=off} % DO NOT CHANGE THIS
\floatstyle{ruled}
\newfloat{listing}{tb}{lst}{}
\floatname{listing}{Listing}
\usepackage{amsmath}
\usepackage{graphicx}

\usepackage{multirow}
\usepackage[export]{adjustbox}
\usepackage{mathrsfs}
\usepackage{array}

\title{Blend the Separated: Mixture of Synergistic Experts for Data-Scarcity Drug-Target Interaction Prediction}
\author{
    Xinlong Zhai\textsuperscript{\rm 1},
    Chunchen Wang\textsuperscript{\rm 1},
    Ruijia Wang\textsuperscript{\rm 2},
    Jiazheng Kang\textsuperscript{\rm 1},
    Shujie Li\textsuperscript{\rm 1},\\
    Boyu Chen\textsuperscript{\rm 1},
    Tengfei Ma\textsuperscript{\rm 3},
    Zikai Zhou\textsuperscript{\rm 1},
    Cheng Yang\textsuperscript{\rm 1},
    Chuan Shi\textsuperscript{\rm 1}\thanks{Corresponding author},    
}
\affiliations{
    \textsuperscript{\rm 1}Beijing University of Posts and Telecommunications\\
    \textsuperscript{\rm 2}China Telecom Cloud Computing Research Institute\\
    \textsuperscript{\rm 3}College of Computer Science and Electronic Engineering, Hunan University\\
    \{zhaijojo, wangchunchen, kjz, shujielie, chenbys4, 2020211668, yangcheng, shichuan\}@bupt.edu.cn,\\
    wangrj12@chinatelecom.cn, tfma@hnu.edu.cn
}

\usepackage{bibentry}
\begin{document}

\maketitle

\begin{abstract}
Drug-target interaction prediction (DTI) is essential in various applications including drug discovery and clinical application. There are two perspectives of input data widely used in DTI prediction: Intrinsic data represents how drugs or targets are constructed, and extrinsic data represents how drugs or targets are related to other biological entities. However, any of the two perspectives of input data can be scarce for some drugs or targets, especially for those unpopular or newly discovered. Furthermore, ground-truth labels for specific interaction types can also be scarce. Therefore, we propose the first method to tackle DTI prediction under input data and/or label scarcity. To make our model functional when only one perspective of input data is available, we design two separate experts to process intrinsic and extrinsic data respectively and fuse them adaptively according to different samples. Furthermore, to make the two perspectives complement each other and remedy label scarcity, two experts synergize with each other in a mutually supervised way to exploit the enormous unlabeled data. Extensive experiments on 3 real-world datasets under different extents of input data scarcity and/or label scarcity demonstrate our model outperforms states of the art significantly and steadily, with a maximum improvement of 53.53\%. We also test our model without any data scarcity and it still outperforms current methods. 

\end{abstract}

\begin{links}
\link{Code}{https://github.com/BUPT-GAMMA/MoseDTI}
\end{links}

\section{Introduction}
\label{sec_intro}
The task of drug-target interaction (DTI) prediction is crucial across various biological fields, particularly within the pharmacology \cite{lukavcivsin2019emergent, bredel2004chemogenomics, lee2019computational, KGE-UNIT}. In this task, a drug (molecule) and a target (a gene or the encoded protein of a gene) are input and output is the probability of them interacting.

\begin{figure}[t]
    \centering
    \includegraphics[width=\linewidth]{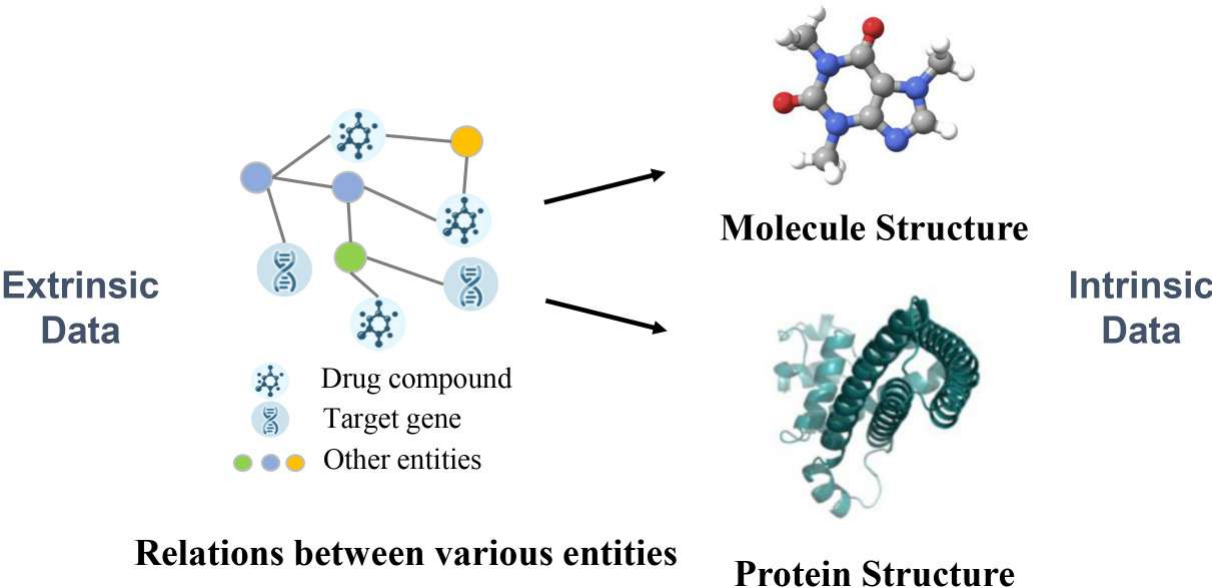}
    \caption{Illustration of intrinsic and extrinsic data.}
    \label{fig_intro}
\end{figure}

There has been a surge in the development of diverse neural networks for DTI prediction, which significantly reduces the need for domain knowledge and has demonstrated superior results. Generally, there are two perspectives of data which can be utilized in these methods, which are illustrated in Fig. \ref{fig_intro}. The first perspective of data is how molecules or proteins are composed, like molecule structures and amino acid residue sequences of proteins. We denote this perspective of data as intrinsic data. DeepDTA \cite{ozturk2018deepdta} uses two separate CNNs to encode the SMILES representation of molecule structures and amino acid residue sequences of proteins respectively. The second perspective of data is the relations between various biological entities, such as diseases, side effects and symptoms, besides the drugs and targets. We denote this perspective of data as extrinsic data. Entities and relations can be formed as graphs so various graph embedding methods can be applied \cite{su2024amgdti, wang2022heterogeneous}. It is natural to consider utilizing both perspectives of data to achieve better prediction performance, and more recently, there have emerged a few methods to realise it. MDTips \cite{xia2023mdtips} uses a ConvE to embed extrinsic data, a GAT and a transformer to embed intrinsic data of drugs and targets, and then concatenate them to predict interaction.

However, there are two forms of data scarcity that limit the usage of all current works: (1) Intrinsic or extrinsic input data scarcity. With regard to intrinsic data, for example, the acquisition of the most accurate and precise structure of proteins still relies on wet experiments with expensive equipment like cryo-electron microscopes, causing the scarcity of precise protein structures. For extrinsic data, though there has been massive relation data between biology entities accumulated, newly discovered or unpopular drugs or targets could still have few connections with other entities. (2) Interaction label scarcity. The interactions between drugs and targets have diverse specific types. Though there are abundant binary labels of whether they interact, the labels for a specific interaction type are still limited. For example, though there are about 210k DTI labels in the DRKG \cite{drkg2020} dataset, some specific interaction types, like "positive allosteric modulator", could only have dozens of labels, which are insufficient for common deep learning methods.

The main research goal of this work is to propose a method that exploits both intrinsic and extrinsic data effectively, while still functional under input data and/or interaction label scarcity. This requires us to address the following two challenges: (1) \textit{How to fuse intrinsic and extrinsic data flexibly and substantially.} Models with a direct fusion of intrinsic and extrinsic data, like concatenating embeddings from two perspectives, cannot predict when one perspective of data is absent. Furthermore, when predicting without one data perspective, how could we still utilize the knowledge learnt from data of this perspective during training? (2) \textit{How to optimize efficiently with limited interaction labels.} Caused by the huge divergence between different specific relations, we cannot transfer the knowledge learnt in the general interaction to specific interactions to remedy the interaction label scarcity, which is demonstrated experimentally in Appendix B. Moreover, intrinsic data contains composition information of the drugs and targets themselves while extrinsic data contains higher-level semantic information between drugs and targets. Hence it also remains to be explored how to optimize models more label-efficiently by exploiting the complementarity between the two data perspectives.

In this paper, we propose a novel method MoseDTI, i.e., \textbf{m}ixture \textbf{o}f \textbf{s}ynergistic \textbf{e}xperts for data-scarcity \textbf{d}rug-\textbf{t}arget \textbf{i}nteraction prediction, which performs well under any or both of these two types of data scarcity. We propose a novel model architecture called the mixture of synergistic experts to address the two challenges unitedly and organically. To address the first challenge, two heterogeneous experts are designed to predict DTI interaction probabilities according to intrinsic and extrinsic data respectively. Then a gating model is applied to adaptively adopt their output according to whether the intrinsic or extrinsic data of a sample is more reliable. The two experts are synergized, i.e., one expert supervises the other during training to inject knowledge from one perspective into the other expert. If intrinsic or extrinsic data is absent when predicting, one of the experts can still predict normally. To address the second challenge, the two experts are designed to generate pseudo labels for each other as the supervision method. The pseudo labels generated effectively enlarge the training samples for the two experts and the gating model, and adequately exploit the complementarity between the two perspectives of data.

Elaborate experiments on three real-world datasets under any or both data scarcity of different extents show that our method outperforms state-of-the-art steadily and significantly, with a maximum improvement of 53.53\%. We also test our method on two real-world datasets without data scarcity and it still outperforms other methods, which proves the generality of our method.

\section{Related Work}
In this section, we classify all current DTI works into intrinsic methods, extrinsic methods and hybrid methods according to which data perspective they use, which is elaborated in Sec. \ref{sec_intro}, and roughly review them.

\textbf{Intrinsic methods.}
Many works utilizing various deep neural networks have achieved excellent performance for drug-target interaction prediction to encoder intrinsic data of drugs and targets \cite{tsubaki2019compound_protein, 10.1007/978-3-030-45257-5_29, chen2020transformercpi, chen2023sequence, nguyen2021graphdta, ozturk2018deepdta, karimi2019deepaffinity}. An end-to-end deep learning framework named GNN-CPI \cite{tsubaki2019compound_protein} that applied the GNN to embed the compound represented by molecular graph is an early work. MONN \cite{10.1007/978-3-030-45257-5_29} was proposed to jointly predict both non-covalent interactions and binding affinities between compounds and proteins. TransformerCPI2.0 \cite{chen2023sequence} introduces a sequence-to-drug concept, employing end-to-end differentiable learning based on protein sequences.  These methods only use the local features of drugs and targets themselves ignoring there are abundant extrinsic data between biology entities and cannot predict a specific interaction type with few-shot labels. From another perspective, they can also be seen as orthogonal to our work because the drug and target encoder in our model can be easily replaced by the encoders presented in these works.

\textbf{Extrinsic methods.}
Some studies on DTI prediction apply extrinsic data and resolve a link prediction task on a graph or a heterogeneous information network \cite{mohamed2019discovering, su2024amgdti, ezzat2016drug, wan2019neodti, peng2021end, wang2022heterogeneous, li2021imchgan}. For example, TriModel \cite{mohamed2019discovering} adopted KG embedding to learn the representations of drugs and targets for DTI prediction. AMGDTI \cite{su2024amgdti} introduces an adaptive meta-graph learning approach and automates semantic information aggregation from heterogeneous networks for DTI prediction. However, these works ignore the intrinsic data, i.e., the local features of nodes themselves, which is fatal when we need to predict newly discovered or unpopular drugs or targets with few connections to other biological entities.

\textbf{Hybrid methods.}
We also noticed that there are attempts to utilize both intrinsic and extrinsic data for better DTI prediction performance \cite{zhou2021multidti, ma2022kg, xia2023mdtips, li2023sagdti, dong2023multi}. KG-MTL \cite{ma2022kg} tries to merge knowledge graph and molecule graph via multi-task learning, which employs a shared unit to jointly maintain drug entity semantics and compound structural relations in both graphs. MDTips \cite{xia2023mdtips} predicts DTI using multi-modal data, integrating knowledge graphs, gene expression profiles, and structural details of drugs and targets. However, They are not designed to handle DTI prediction with limited labels and are hard to predict when extrinsic or intrinsic data is absent for some samples.

\section{Preliminaries}

\textbf{Extrinsic data.} We consider extrinsic data as a knowledge graph (KG) as $\mathcal{KG} = (E, R, O)$ that provides abundant relation information between different kinds of biological entities, where E is a set of entities, R is a set of relations, and O is a set of observed $(\mathsf{h}, \mathsf{r}, \mathsf{t})$ triples. In a triple, $\mathsf{h}, \mathsf{r}, \mathsf{t}$ represents the head entity, relation, and tail entity respectively. The entity set $E$ contains various biological entities such as diseases, side-effects and symptoms, and the drug and target sets are subsets of the entity set: $D, T \subset E$. To prevent label data leakage, we remove all direct connections of drugs and targets from $\mathcal{KG}$, i.e. remove all $(\mathsf{h}, \mathsf{r}, \mathsf{t})$ from $O$ which satisfies $\mathsf{h} \in D, \mathsf{t}\in T$ or $\mathsf{h} \in T, \mathsf{t}\in D$.

\textbf{Intrinsic Data.} For a drug $d_i \in D$, we use the SMILES sequence $SM_{d_i}$, i.e., \textit{simplified molecular-input line-entry system} sequence as its intrinsic feature, which is a specification in the form of a line notation for describing the structure of chemical substance using short ASCII strings. For a target gene $t_i \in T$, we use the UniProt database to obtain the amino acid sequence of the protein it encodes as its intrinsic feature, denoted as $AS_{t_i}$.

\textbf{DTI Task.} In the drug-target interaction task, we aim to estimate the interaction probability $p_{ij}$ of a drug-target pair $(d_i, t_j)$ under a specific interaction type, where $d_i \in D, t_j \in T$. Such a DTI dataset can be described as $(X^{p}, X^{n}, \mathcal{KG}, SM_{D}, AS_{T})$, where $X^{p}$ or $X^{n}$ is $\{(d_i, t_j)\}$ which indicates these drug-target pairs have or do not have this type of interaction, and $SM_{D}$ and $AS_{T}$ denote intrinsic data of all drugs and targets respectively.

\section{Methodology}
\begin{figure*}
    \centering
    \includegraphics[width=0.8\linewidth]{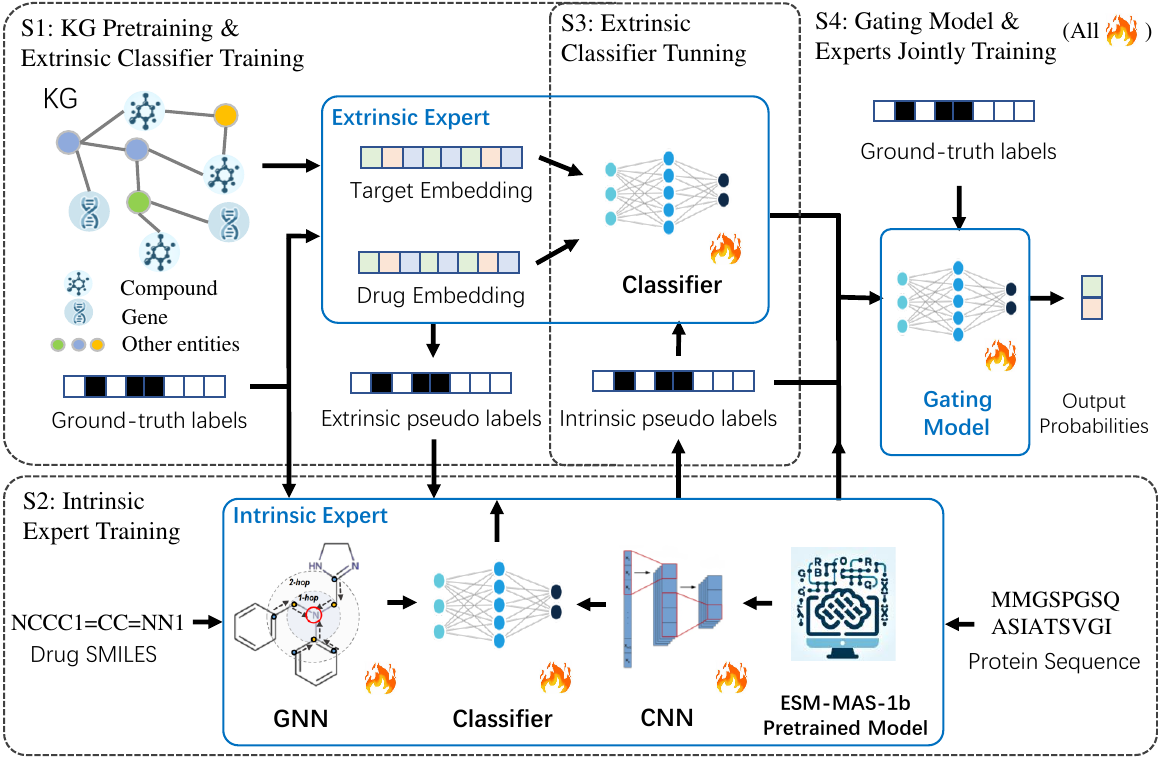}
    \caption{The framework of our MoseDTI. The three components are surrounded in blue rectangles and the first three training steps S1 to S3 are surrounded in dotted black rectangles. For the last step S4, all components with little flames are jointly trained.}
    \label{fig_framework}
\end{figure*}

In this section, we describe the proposed MoseDTI model for drug-target interaction prediction with data scarcity specifically, as illustrated in Fig. \ref{fig_framework}. We first introduce the model structure of two heterogeneous experts and a gating model, and then elaborate on how we optimize them. 

\subsection{Model Architecture}
Our model consists of three components: an extrinsic expert, an intrinsic expert and a gating model. The two experts take extrinsic and intrinsic data as input respectively and output interaction probabilities. The gating model takes intrinsic and extrinsic representations of both the drug and target and output weight to determine whether the intrinsic or extrinsic expert is more reliable for the current sample. This architecture utilizes the extrinsic and intrinsic data adaptively according to specific samples, and the two experts can work alone if one perspective of data is absent when predicting.

\subsubsection{Extrinsic Expert}
The extrinsic expert is to predict DTI based on relation data between biological entities. We first use the massive unlabeled association data between biological entities to pretrain embeddings of drugs and targets, and then train a classifier with labels to output the interaction probabilities from the extrinsic perspective.

\textbf{Knowledge graph embedding.} The knowledge graph $\mathcal{KG}$ contains various association data between different biological entities, in which drugs and targets are connected to other types of entities, like diseases, side effects, and symptoms. To leverage the abundant semantic information it implies, we first use the KG embedding method to pretrain the $d$-dimensional drug extrinsic embedding $\mathbf{h}^{ex}_{d_i} \in \mathcal{R}^d$ for drug $d_i$ and target extrinsic embedding $\mathbf{h}^{ex}_{t_j} \in \mathcal{R}^d$ for target $t_j$. We do not introduce any labelled drug-target interaction data into the pretrain, so the embeddings can be used for different specific DTI datasets without retraining.

\textbf{Extrinsic classifier.} After that, given a specific interaction dataset  and its ground-truth samples $(X^{p}, X^{n})$, there is an extrinsic classifier $g_r$ to predict the interaction probability for $(d_i, t_j)$:

\begin{equation}
    \label{equ_pi_ex}
    p^{ex}_{ij} = g^{ex}(\mathbf{h}^{ex}_{d_i}, \mathbf{h}^{ex}_{t_j})
\end{equation}

In the experiment, we implement the $g_r$ as a simple MLP because a simpler $g_r$ with fewer parameters can be more easily trained by limited samples.

\subsubsection{Intrinsic Expert}
The intrinsic expert is to predict DTI based on the structure data of drugs and targets. We use a drug encoder and a target encoder to encode drug SMILES sequence and target amino acid residue sequences respectively. Then, an intrinsic classifier is applied to output the interaction probability intrinsically.

\textbf{Drug encoder.} For a drug $d_i$, its SMILES sequence $SM_{d_i}$ is first translated to a molecule graph $\mathcal{MG}_{d_i}$ with RDKit \cite{landrum2006rdkit}. $\mathcal{MG}_{d_i}=(\mathcal{V}_{d_i},\mathcal{E}_{d_i})$, where $\mathcal{V}_{d_i}$ denotes the set of nodes, i,e., atoms and $\mathcal{E}_{d_i}$ is the set of edges between atoms, i.e., chemical bonds. A node $v \in \mathcal{V}_{d_i}$ has its embedding initialized as $\mathbf{h}_v^{(0)}$ with the method proposed in \cite{quan2019graphcpi}.  We utilize a graph neural network (GNN) to obtain the final embedding of each node \cite{gilmer2017neural, quan2019graphcpi}:

\begin{align}
\mathbf{m}_{u \rightarrow v}^{(l)}&=Message^{(l)}\left(\mathbf{h}_v^{(l-1)}, \mathbf{h}_v^{(l-1)}\right) \\
\mathbf{m}_v^{(l)}&=Reduce_{u \in \mathcal{N}(v)} \mathbf{m}_{u \rightarrow v}^{(l)} \\
\mathbf{h}_v^{(l)}&=Update^{(l)}\left(\mathbf{h}_v^{(l-1)}, \mathbf{m}_v^{(l)}\right),
\end{align}

where $Message$, $Reduce$ and $Update$ are three functions specified by the selected GNN, like GCN \cite{Kipf2017SemiSupervisedCW} or GAT \cite{velickovic2017graph}. The superscript $(l), l=1,2,...,L$ indicates a certain GNN layer, and the $\mathcal{N}(v)$ denotes the nodes connected to $v$ by edges. $\mathbf{m}_{u \rightarrow v}^{(l)}$  and $\mathbf{m}_v^{(l)}$ indicate the message from $u$ to $v$  and the overall message $v$ received at layer $(l)$ respectively.  Then, the embedding of the whole molecule graph $\mathbf{h}^{in}_{d_i}$ is calculated by a multi-layer perceptron (MLP) from the $L-th$ layer embeddings and a max readout function:
\begin{equation}
    \mathbf{h}^{in}_{d_i} = max(\{MLP(\mathbf{h}_v^{(L)})|v \in V\}).
\end{equation}

\textbf{Target encoder.} For a target $t_j$, its amino acid residues sequence $AS_{t_j}$ is first input to a pre-trained protein language model ESM-MSA-1b \cite{rives2021biological} and embeddings $\{\mathbf{e}_m^{(0)} | m= 1,...,M\}$for every amino acid residue are output, where $M$ is the length of the sequence. Then, a $K$ layer 1-dimensional CNN with adaptive max pooling is applied to obtain the final embedding for the sequence \cite{ma2022kg}:

\begin{equation}
\mathbf{h}_{t_j}^{in} = AMP(Conv(
\{\mathbf{e}_j^{(0)} | k=1,...,K\})),
\end{equation}

where $Conv = Conv^{(1)} \circ \cdots \circ Conv^{(K)}$ and each convolution layer contains a 1-dimensional convolution operation and a ReLU activation function, and $AMP$ represents adaptive max pooling.

\textbf{Intrinsic classifier.}
Subsequently, given a specific interaction dataset, an intrinsic classifier $g^{in}$ is trained to predict the interaction probability:

\begin{equation}
    p^{in}_{ij} = g^{in}(\mathbf{h}^{in}_{d_i}, \mathbf{h}^{in}_{t_j}).
\end{equation}

Experimentally, like the extrinsic classifier $g^{ex}$, we also implement $g^{in}$ as a MLP.

\subsubsection{Gating model} To exploit both extrinsic and intrinsic adaptively, we fuse the output of two experts according to specific samples. We design a gating model \cite{jacobs1991adaptive} that accepts the hidden embeddings of the two experts to generate a weight $w_{ij}$:
\begin{equation}
    w_{ij} = Gating(\mathbf{h}^{ex}_{d_i},\mathbf{h}^{ex}_{t_j},\mathbf{h}^{in}_{d_i},\mathbf{h}^{in}_{t_j}).
\end{equation}

We implement $Gating$ as an MLP and a softmax function. Then we use $w_{ij}$ to blend the two models as the final output of our entire model:
\begin{equation}
    \label{equ_pi}
    p_{ij} = w_{ij} p_{ij}^{ex} + (1-w_{ij})p_{ij}^{in}.
\end{equation}

\subsection{Optimization}
 In this subsection, we first introduce how experts are synergized and then elaborate on the entire training procedures.

\subsubsection{Synergizing experts} Inspired by the self-training optimization strategy, we design a novel expert synergizing mechanism, in which experts generate pseudo labels from massive unlabeled samples for each other. Compared to the self-training, the exchange of pseudo labels could bring in high-confidence samples from another perspective, preventing the model from over-fitting to the samples similar to the offered ground-truth samples. Here we elaborate on how one expert called $ExpertA$ generates pseudo labels with two-stage sampling to train another expert $ExpertB$.

First, we sample a portion of the cartesian set of the drug set $D$ and the target set $T$ as a candidate set:
\begin{equation}
    Cand_A = Sample(D \times T, \lfloor \alpha_A L_{ca} \rfloor),
\end{equation}
where $\times$ denotes the cartesian product of two sets, $L_{ca} = |D||T|$ is length of the cartesian product set and $\alpha_A$ is the sampling rate . The length of $Cand_A$ is denoted as $L_{Cand_A} = \lfloor \alpha_A L_{ca} \rfloor$. Next, we use expert A to predict on $Cand_A$:
\begin{equation}
    P_{Cand_A} = ExpertA(Cand_A),
\end{equation}
where $P_{Cand_A} = \{p^A_s | s=1,...,L_{Cand_A}\}$ and $p^A_s$ is the probability of interaction predicted by $ExpertA$ for drug-target pair $s$. Then we sort $Cand_A$ according to $P_{Cand_A}$ and select top samples in $Cand_A$ as pseudo positive samples, denoted as $X_A^{p}$:
\begin{equation}
\label{equ_select_top}
    X_A^{p} = Top(Cand_A, P_{Cand_A}, \lfloor\beta_A L_{Cand_A}\rfloor),
\end{equation}
where $X_A^{p} = \{(d_i, t_j)\}$,  its length $L_{X_A^{p}} = \lfloor \beta_A L_{Cand_A} \rfloor$ and $\beta_A$ is a choosing rate.  Denoting the ground-truth positive and negative samples as $X^{p}$   and $X^{n}$, the expert B is trained with loss:
\begin{multline}
    \label{equ_syn_loss}
    \mathcal{L}_B = - \left( \sum_{(d_i,t_j) \in \gamma_A X^{p} \cup X_A^{p} }\log(p_{ij}^B) +\right.
    \\ \left.\sum_{(d_i,t_j) \in X^{n} \cup X_A^{n} } \log(1 - p_{ij}^B)\right)
\end{multline},
where $X_A^{n}$  is pseudo negative samples and $\gamma_A$ is an integer to amplify the weight of true positive samples versus pseudo positive
samples. Considering that most samples in $Cand_A$ are actually negative samples, to improve the diversity of negative samples, we select the bottom samples from $Cand_A$ according to $P_{Cand_A}$ with a larger length $|X_A^{n}|$ which satisfies
\begin{equation}
|X_A^{n}| = \gamma_A |X^{p}| + |X_A^{p}| - |X^{n}|
\end{equation}
to keep the label balanced.

\subsubsection{Training procedures}
Due to the scarcity of interaction labels,  in the first step S1 in Fig. \ref{fig_framework}, we first pretrain the KG embedding with classical methods like TransE \cite{bordes2013translating} or RotatE \cite{sun2019rotate}, exploiting all the observed triples in the knowledge graph. Then we only train the extrinsic classifier in Equ.\ref{equ_pi_ex} with relatively much fewer parameters above the frozen pre-trained embeddings, using the ground-truth labels:
\begin{equation}
    \mathcal{L}_{s1} = -\left(\sum_{(d_i,t_j)\in X^{p}} \log(p_{ij}^{ex}) + \sum_{(d_i,t_j)\in X^{n}} \log(1 - p_{ij}^{ex})\right).
\end{equation}
In the second step S2, the extrinsic expert is the $ExpertA$, the intrinsic expert is the $ExpertB$ and the intrinsic expert is trained with Equ.\ref{equ_syn_loss}.  In the third step S3, the two models exchange their positions, and the extrinsic expert is tuned with Equ. \ref{equ_syn_loss}. In the last step S4, we jointly train the gating model, intrinsic, and extrinsic expert with ground-truth and pseudo labels from two experts:
\begin{multline}
    \mathcal{L}_g = -\left(\sum_{(d_i,t_j)\in \gamma_g X^{p}\cup X_A^{pos'} \cup X_B^{pos'}} \log(p_{ij}) + \right. \\
    \left. \sum_{(d_i,t_j)\in X^{n}\cup X_A^{neg'} \cup X_B^{neg'}} \log(1 - p_{ij})\right).
\end{multline}
Similar to Equ.\ref{equ_select_top}, the positive pseudo samples $X_A^{pos'}$ and $X_B^{pos'}$ are also selected from the top of $Cand_A$ and $Cand_B$ with a shared rate $\beta_{g}$ and tailing samples of the longer one are trimmed to keep their length equal. $X_A^{neg'}$ and $X_B^{neg'}$ are also selected from the bottom of $Cand_A$ and $Cand_B$ according to $P_{Cand_A}$ and $P_{Cand_B}$ respectively, with their lengths
\begin{equation}
    |X_A^{neg'}| = |X_B^{neg'}| = (\gamma_g |X^{p}|+ |X_A^{p}| + |X_B^{p}| - |X^{n}|) / 2,
\end{equation}
to balance the total positive and negative samples.

\section{Experiments}

\begin{table*}[t]

\begin{adjustbox}{width=\textwidth}
\renewcommand{\arraystretch}{1.3}
{\Huge
\begin{tabular}{llccccccccc}
\hline
                                                      &                               & \multicolumn{3}{c}{DGIDB::AGONIST}                                 & \multicolumn{3}{c}{DGIDB::BLOCKER}                                 & \multicolumn{3}{c}{GNBR::E-}                                       \\
                                                      &                               & ACC                  & AUC                  & AUPR                 & ACC                  & AUC                  & AUPR                 & ACC                  & AUC                  & AUPR                 \\ \hline
\multicolumn{1}{c}{\multirow{4}{*}{Intrinsic Method}} & \multicolumn{1}{c}{GNNCPI}    & 51.10±1.03           & 61.22±2.77           & 61.89±1.85           & 57.63±3.63           & 70.23±5.73           & 73.51±5.49           & 54.39±3.29           & 64.71±3.84           & 65.81±3.44           \\
\multicolumn{1}{c}{}                                  & \multicolumn{1}{c}{TFCPI}     & 58.55±3.42           & 66.11±6.33           & 66.37±4.67           & 56.27±5.07           & 68.15±1.72           & 63.94±3.05           & 63.10±3.20           & 78.14±0.37           & 78.06±0.45           \\
\multicolumn{1}{c}{}                                  & \multicolumn{1}{c}{TFCPI2.0}  & 50.79±0.10           & 53.05±0.11           & 50.77±0.12           & 36.66±0.25           & 33.53±0.42           & 38.85±0.16           & 42.07±0.07           & 40.49±0.12           & 41.90±0.06           \\
\multicolumn{1}{c}{}                                  & \multicolumn{1}{c}{Mose-intr} & 64.50±6.31           & 73.70±6.40           & 72.47±6.91           & 88.48±4.64           & 93.17±3.64           & 94.34±2.34           & 70.64±4.32           & 80.43±3.75           & 78.83±4.32           \\ \hline
\multicolumn{1}{c}{\multirow{5}{*}{Extrinsic Method}} & \multicolumn{1}{c}{TransE}    & 50.00±0.00           & 50.22±0.08           & 50.65±0.08           & 50.00±0.00           & 53.96±0.23           & 55.22±0.48           & 50.00±0.00           & 48.59±0.04           & 48.62±0.06           \\
\multicolumn{1}{c}{}                                  & \multicolumn{1}{c}{RotatE}    & 50.00±0.00           & 50.41±0.07           & 50.49±0.05           & 50.00±0.00           & 51.24±0.40           & 50.98±0.52           & 50.00±0.00           & 48.84±0.07           & 49.91±0.08           \\
\multicolumn{1}{c}{}                                  & \multicolumn{1}{c}{TriModel}  & 50.19±0.15           & 48.34±1.23           & 50.21±1.29           & 50.00±0.00           & 33.85±5.38           & 42.09±2.57           & 50.01±0.04           & 49.59±1.81           & 50.08±0.98           \\
\multicolumn{1}{c}{}                                  & \multicolumn{1}{c}{AMGDTI}    & 57.91±3.72           & 66.48±2.87           & 69.42±6.84           & 80.89±2.42           & 96.14±2.35           & 96.08±0.36           & 61.49±8.87           & 63.84±1.24           & 64.28±1.69           \\
\multicolumn{1}{c}{}                                  & \multicolumn{1}{c}{Mose-extr} & 64.88±3.54           & 75.55±3.94           & 75.41±4.48           & 80.28±2.18           & 98.11±0.29           & 97.14±0.31           & 70.87±4.08           & 88.03±4.52           & \textbf{87.24±4.74}  \\ \hline
\multicolumn{1}{c}{\multirow{3}{*}{Hybrid Method}}    & \multicolumn{1}{c}{KG-MTL}    & 56.76±1.02           & 56.06±2.71           & 50.22±0.89           & 65.83±5.69           & 73.57±4.09           & 76.07±3.16           & 54.06±0.71           & 57.65±1.08           & 57.61±1.16           \\
\multicolumn{1}{c}{}                                  & \multicolumn{1}{c}{MDTips}    & 64.23±4.13           & 73.51±2.91           & 72.37±2.87           & 91.25±2.34           & 97.27±0.86           & 96.90±0.79           & 69.61±3.43           & 82.35±3.89           & 80.78±4.86           \\
\multicolumn{1}{c}{}                                  & \multicolumn{1}{c}{MoseDTI}   & \textbf{67.27±1.90}  & \textbf{75.82±2.90}  & \textbf{75.57±3.62}  & \textbf{92.85±2.79}  & \textbf{98.76±0.82}  & \textbf{97.21±0.64}  & \textbf{78.17±4.86}  & \textbf{88.71±4.34}  & 86.62±5.31           \\ \hline
                                                      
\end{tabular}
}
\end{adjustbox}
  \caption{Model performance on three 10-shot datasets of specific interaction comparing three variants of our model including MoseDTI, Mose-intr and Mose-extr with nine baselines. The best performance is boldfaced.}
\label{tab_few_shot}
\end{table*}

In this section, we first introduce on the datasets and baselines and then show model results.  The goal of our experiments is to answer the following research questions (RQs). 

\begin{enumerate}
    \item  Can MoseDTI effectively confront input data scarcity (including intrinsic or extrinsic data scarcity) and/or interaction label scarcity? (\textbf{RQ1})
    \item  If there is no data scarcity, can MoseDTI still perform well? (\textbf{RQ2})
    \item  Are the MOE architecture and the synergizing mechanism beneficial? (\textbf{RQ3})
\end{enumerate}

We also conduct experiments of hyper-parameters, case studies and few-shot learning on general interaction datasets. Please see Appendix F, G and H respectively.

\subsection{Experimental Setup}

\subsubsection{Datasets} We conduct experiments on 5 datasets. There are 3 few-shot datasets of specific interactions including DGIDB::BLOCKER (blocker), DGIDB::AGONIST (agonist) \cite{griffith2013dgidb} and GNBR::E- (e-) \cite{percha2018global}. Each dataset presents a specific interaction type and only contains 10 positive ground-truth samples for training. There are also two normal datasets of general DTI interaction including DrugBank \cite{wishart2018drugbank} and DrugCentral \cite{10.1093/nar/gkw993}, which contain 18480 and 18066 samples respectively and are partitioned into train, valid and test with a ratio of 6:2:2. All of the datasets use the DRKG \cite{drkg2020} as the common extrinsic data and all connections between drugs and targets are removed to prevent data leakage. The SMILES of drugs are also from DrugBank \cite{10.1093/nar/gkw993}. We collect the amino acid residue sequences of proteins coded by targets from UniProt\footnote{https://www.uniprot.org/}. More details of datasets and the evaluation protocol are in Appendix C and D.

\subsubsection{Baselines}
We use 8 baselines and classify them into intrinsic methods, extrinsic methods and hybrid methods according to whether they only use the intrinsic or extrinsic data, or use both perspectives of data, which is elaborated in Sec. \ref{sec_intro}. Intrinsic methods include GNNCPI \cite{tsubaki2019compound_protein}, TransformerCPI \cite{chen2020transformercpi} and TransformerCPI2.0 \cite{chen2023sequence}. Extrinsic methods include TransE \cite{bordes2013translating}, RotatE \cite{sun2019rotate}, TriModel \cite{mohamed2019discovering}, AMGDTI \cite{su2024amgdti}. Hybrid methods include KG-MTL \cite{ma2022kg} and MDTips \cite{xia2023mdtips}. For the implementation details of our model and baselines, see Appendix E.

\subsection{Performance under Data Scarcity (RQ1)}
\label{sec_data_scarcity}

To validate that MoseDTI is effective under different scenarios of data scarcity, we conduct exhaustive experiments with the following two orthogonal scarcity settings:
(1) For intrinsic or extrinsic data scarcity, there are 3 different settings of data availability when inference: only intrinsic data is available; only extrinsic data is available; both perspectives of data are available. We take the intrinsic expert in our method as Mose-intr, the extrinsic expert as Mose-extr, after the entire training procedure, which can predict when only one data perspective is available.
(2) For interaction label scarcity, there are also three scarcity extents regarding to the number of labelled positive samples for training, i.e., 10-shots, 20-shots and 40-shots. 

we compare MoseDTI and its variants with state-of-the-art methods on three real-world datasets under totally 9 ($3 \times 3$) differnet scarcity settings. The results of 10-shots are shown in Tab. \ref{tab_few_shot}. The results of 20 and 40 shots are shown in Appendix I. Generally, our method significantly outperforms other methods.

The first four methods can be applied when intrinsic data is absent and only extrinsic data is available
 for prediction. GNN-CPI, which applies a GNN to encode molecular graphs of compounds and a CNN to obtain
  chemical features of proteins, demonstrates its stable but limited ability to confront few-shot settings.
   TransformerCPI2.0 (TFCPI2.0) claims its excellent performance for generalizing to new compounds and
    proteins, while it fails to predict specific interaction types, probably owing to the substantial difference between specific and general DTI interaction. Our Mose-intr makes use of a pre-trained protein language model and massive unlabeled drug-target pairs when training, outperforming them steadily by a large margin.

The following five methods can be applied when extrinsic data is absent and only extrinsic data is available for prediction. The three KG-embedding-based methods, i.e., TransE, RotatE and TriModel all fail to perform well with limited labelled interactions. It is conceivable that the few-shot labels are not enough for them to optimize their free embeddings of entities and relations associated with a certain interaction type. AMGDTI automatically aggregates semantic information from KG by training an adaptive meta-path and performs pretty well in the blocker dataset. However, our Mose-extr, which also accommodates label scarcity via the designing of the pre-trained entity embedding plus a simple MLP layer and trained with unlabeled potential interaction pairs, performs generally better on the three datasets

The last three methods can be applied when both extrinsic and intrinsic data are available for prediction. KG-MTL fails to perform well, probably because the complex model architecture for handling two related tasks together needs sufficient labelled samples. MDTips fuses the embeddings from KG embeddings, drug embeddings and target embeddings and performs remarkably across three datasets. Thanks to the MOE architecture and the usage of unlabeled samples, our MoseDTI outperforms it obviously and steadily.

\subsection{Performance without Data Scarcity (RQ2)}
\begin{table*}[t]

\begin{adjustbox}{width=\textwidth}

{\Large
\begin{tabular}{cccccccc}
\hline
\multicolumn{1}{l}{}             & \multicolumn{1}{l}{}          & \multicolumn{3}{c}{DrugCentral}                                                            & \multicolumn{3}{c}{DrugBank}                                          \\
\multicolumn{1}{l}{}             & \multicolumn{1}{l}{}          & ACC                   & AUC                   & AUPR                                       & ACC                   & AUC                   & AUPR                  \\ \hline
\multirow{3}{*}{Intrinsic Method} & \multicolumn{1}{c|}{GNNCPI}   & 72.64 ± 0.51          & 78.44 ± 0.12          & \multicolumn{1}{c|}{80.20 ± 0.18}          & 73.82 ± 0.14          & 80.65 ± 0.11          & 81.99 ± 0.10          \\
                                 & \multicolumn{1}{c|}{TFCPI}    & 80.97 ± 1.34          & 88.94 ± 0.46          & \multicolumn{1}{c|}{88.67 ± 0.48}          & 81.50 ± 0.53          & 90.69 ± 0.46          & 90.28 ± 0.39          \\
                                 & \multicolumn{1}{c|}{TFCPI2.0} & 57.58                 & 58.67                 & \multicolumn{1}{c|}{60.60}                 & 54.51                 & 57.19                 & 60.93                 \\ \hline
\multirow{4}{*}{Extrinsic Method} & \multicolumn{1}{c|}{TransE}   & 55.89 ± 0.59          & 74.83 ± 1.02          & \multicolumn{1}{c|}{75.11 ± 1.02}          & 57.86 ± 0.42          & 74.44 ± 0.23          & 76.62 ± 0.46          \\
                                 & \multicolumn{1}{c|}{RotatE}   & 54.34 ± 0.18          & 63.09 ± 0.74          & \multicolumn{1}{c|}{60.22 ± 1.30}          & 60.62 ± 9.72          & 68.17 ± 11.99         & 68.66 ± 12.69         \\
                                 & \multicolumn{1}{c|}{TriModel} & 52.89 ± 0.40          & 63.41 ± 0.60          & \multicolumn{1}{c|}{61.41 ± 0.70}          & 54.25 ± 0.28          & 65.08 ± 0.49          & 64.35 ± 0.87          \\
                                 & \multicolumn{1}{c|}{AMGDTI}   & 80.70 ± 3.32          & 89.27 ± 1.84          & \multicolumn{1}{c|}{89.48 ± 2.70}          & 83.80 ± 0.45          & 90.03 ± 0.62          & 92.57 ± 0.54          \\ \hline
\multirow{3}{*}{Hybrid Method}    & \multicolumn{1}{c|}{KG-MTL}   & 81.60 ± 0.71          & 88.69 ± 0.30          & \multicolumn{1}{c|}{88.98 ± 0.73}          & 80.31 ± 0.46          & 87.50 ± 0.54          & 89.71 ± 0.24          \\
                                 & \multicolumn{1}{c|}{MDTips}   & { 88.10 ± 0.30}    & { 94.62 ± 0.12}    & \multicolumn{1}{c|}{\textbf{95.32 ± 0.21}} & { 87.75 ± 0.58}    & { 94.38 ± 0.15}    & { 94.02 ± 0.23}    \\
                                 & \multicolumn{1}{c|}{MoseDTI}  & \textbf{88.40 ± 0.49} & \textbf{95.11 ± 0.19} & \multicolumn{1}{c|}{\textbf{95.32 ± 0.22}} & \textbf{88.15 ± 0.78} & \textbf{94.90 ± 0.14} & \textbf{95.23 ± 0.28} \\ \hline
\end{tabular}
}
\end{adjustbox}
  \caption{Model performance on general DTI datasets. The best performance is boldfaced.
  }
\label{tab_normal}
\end{table*}

We also test the performance of our model compared with all the baselines above in two datasets of normal label scale for general DTI prediction task. General DTI prediction task regards all kinds of specific interactions as a whole and does not distinguish them, and hence there is abundant labelled interaction data accumulated.

We can observe that GNNCPI, AMGDTI, KG-MTL and TFCPI all perform well, which proves general DTI prediction with plentiful labels is a relatively simple task compared to predicting a specific interaction with a limited amount of labels.  Furthermore, taking advantage of blending two experts adaptively according to the relative importance of intrinsic and extrinsic data of each sample, our MoseDTI even performs better than all other methods.

\begin{figure*}[ht]
    \centering
    \includegraphics[width=0.3\textwidth]{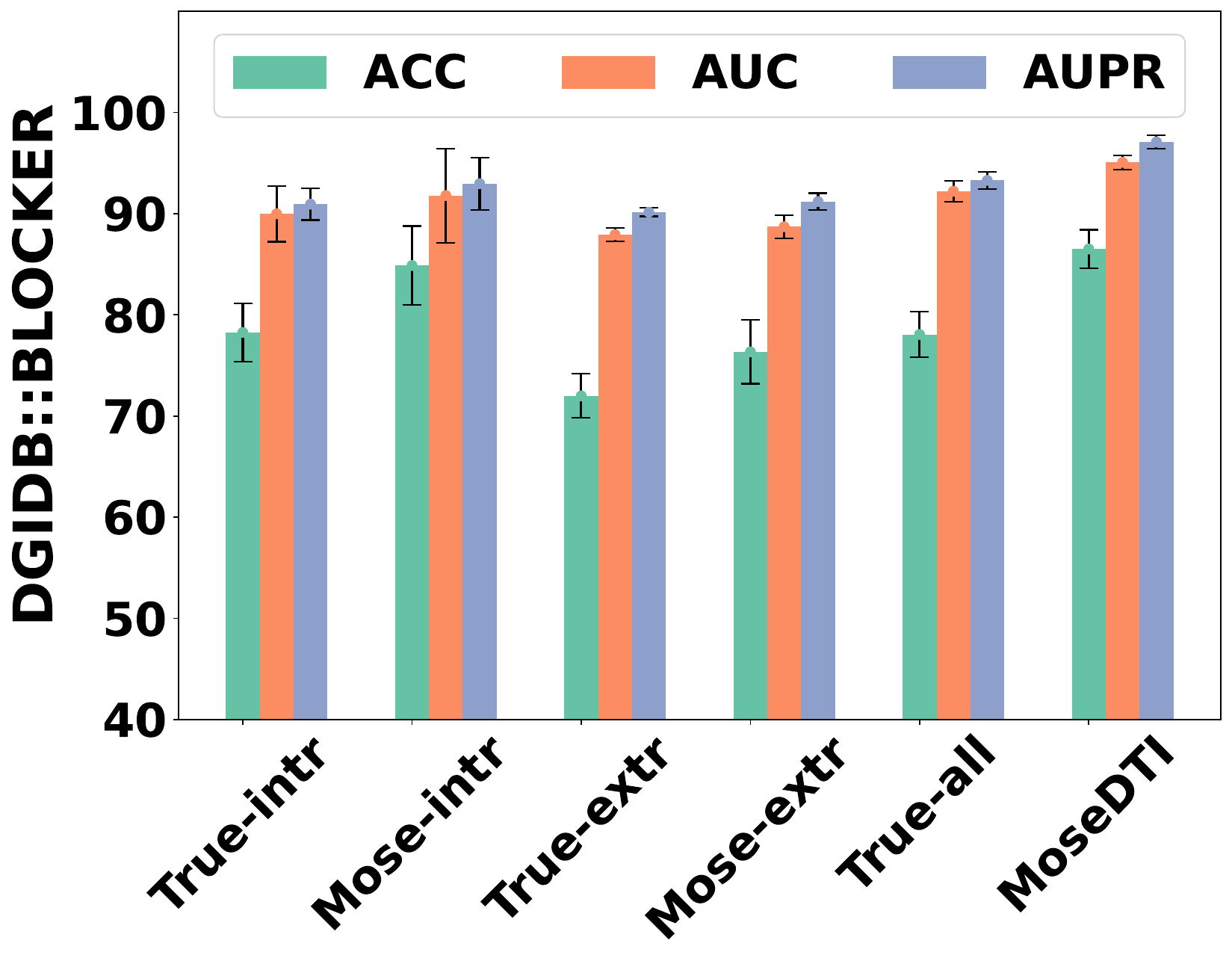}
    \includegraphics[width=0.3\textwidth]{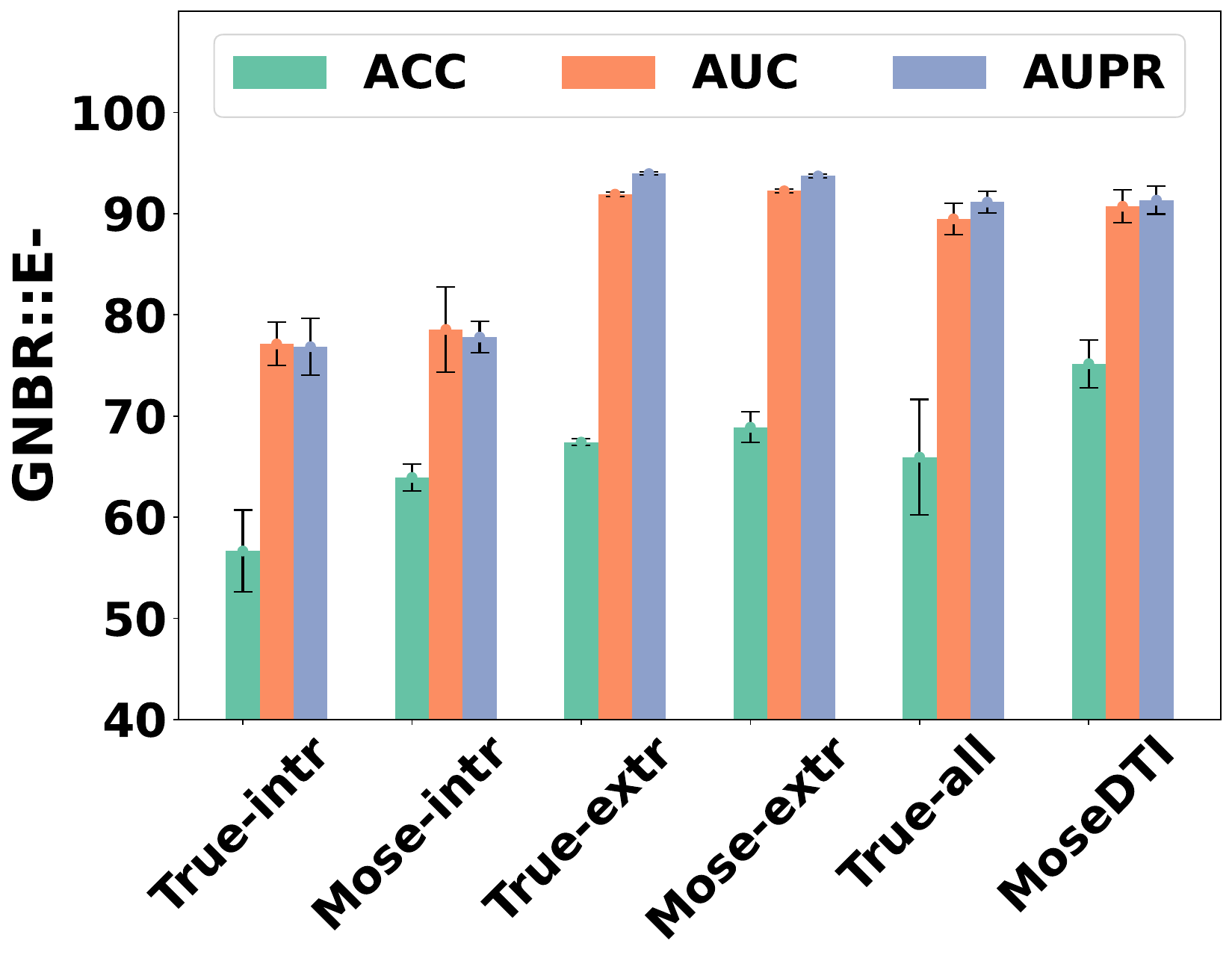}
    \includegraphics[width=0.3\textwidth]{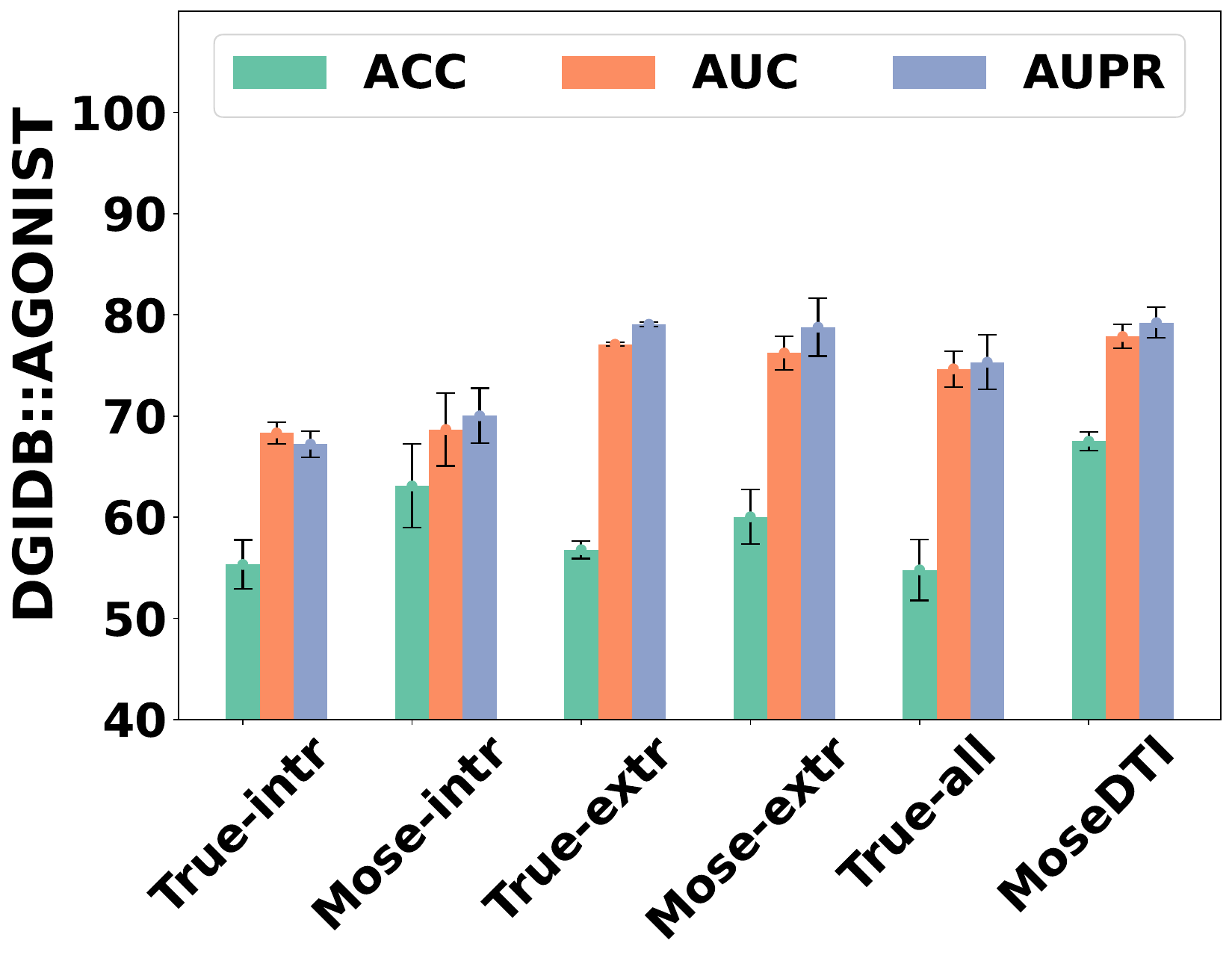}
    \caption{Ablation study on three real-world datasets. The standard deviations are shown in small black lines on top of each bar.}
    \label{fig_ablation}
\end{figure*}

\subsection{Ablation Study (RQ3)}
To investigate how our synergizing mechanism and the design of MOE improve the performance for DTI prediction, we conduct the ablation study with the following variants on few-shot specific DTI prediction: 1) True-intr: training the intrinsic model with only the ground-truth labels. 2) True-extr: training the extrinsic expert with only the ground-truth labels. 3) True-all: training the entire model with only the ground-truth labels. For convenience of comparison, we also add the results of Mose-intr, Mose-extr and MoseDTI to Fig.\ref{fig_ablation}. The difference between True-intr and Mose-intr is whether trained with the synergizing mechanism, and the same with True-extr and Mose-extr. We have the following observations:

 \subsubsection{Effect of synergizing mechanism} The effect of the synergizing mechanism to augment one expert is affected by the performance of the other. For example, on the agonist dataset, the performance of Mose-extr is not obviously better than True-extr, while the improvement is more obvious on the blocker dataset. It is more than likely due to the intrinsic data on the blocker dataset is more easily to be utilized for the intrinsic model to produce pseudo labels with higher quality according to the better performance of True-intr on the blocker dataset compared with the agonist dataset. We also find our synergizing mechanism is robust enough to avoid a negative impact, even if there is a huge gap between the performance of True-intr and True-extr, like on the e- dataset. Furthermore, the performance promotion is even larger when we compare the whole model True-all and MoseDTI on all three datasets, thanks to the synergizing mechanism also providing additional information for the joint training of the whole model.

 \subsubsection{Effect of mixture of experts} The better performance of MoseDTI versus Mose-intr and Mose-extr on all three datasets demonstrates the design of the mixture of experts can enhance both experts while keeping their independence. However, direct training the whole model with few-shot true labels may not surely promote the performance, according to the comparison of True-all against True-intr and True-extr on the e- and agonist datasets.

\section{Conclusion}
This work confronts the problem of effective DTI prediction under input data scarcity (including intrinsic or extrinsic data scarcity) and/or interaction label scarcity. We propose a model architecture: the mixture of synergized experts, which utilizes two synergized heterogeneous experts to process different perspectives of data, which supervise each other mutually with pseudo labels generated from unlabelled samples. The framework solves both forms of data scarcity organically and exhaustive experiments under various data scarcity settings prove its superiority over states of the art.

\textbf{Limitations and Broader Impact.} Despite the encouraging results, there could be more modalities to be incorporated to promote drug-target interaction prediction, such as textual data describing biological entities. Our future work will address these limitations. This research may inspire the AI4Science community to pay more attention to the separation and synergizing of different components when designing their models and make them robust with data scarcity in real-world applications.

\section*{Acknowledgments}
This work is supported by the National Key Research and Development Program of China (2023YFF0725103), the National Natural Science Foundation of China (U22B2038, 62192784), Young Elite Scientists Sponsorship Program (No.2023QNRC001) by CAST and the China Scholarship Council (CSC).

\bibliography{aaai25}

\appendix

\end{document}